# PARSING WITH TYPED FEATURE STRUCTURES

Shuly Wintner

Nissim Francez

Computer Science
Technion, Israel Institute of Technology
32000 Haifa, Israel
{shuly,francez}@cs.technion.ac.il

**Abstract**

In this paper we provide for parsing with respect to grammars expressed in a general TFS-based formalism, a restriction of ALE ([2]). Our motivation being the design of an abstract (WAM-like) machine for the formalism ([14]), we consider parsing as a computational process and use it as an operational semantics to guide the design of the control structures for the abstract machine.

We emphasize the notion of **abstract typed feature structures** (AFSs) that encode the essential information of TFSs and define unification over AFSs rather than over TFSs. We then introduce an explicit construct of **multi-rooted feature structures** (MRSs) that naturally extend TFSs and use them to represent phrasal signs as well as grammar rules. We also employ abstractions of MRSs and give the mathematical foundations needed for manipulating them. We then present a simple bottom-up chart parser as a model for computation: grammars written in the TFS-based formalism are executed by the parser. Finally, we show that the parser is correct.

## 1 Introduction

Typed feature structures (TFSs) serve for the specification of linguistic information in current linguistic formalisms such as HPSG ([10]) or Categorial Grammar ([8]). They can represent lexical items, phrases and rules. Usually, no mechanism for manipulating TFSs (e.g., parsing algorithm) is inherent to the formalism. Current approaches to processing HPSG grammars either translate them to Prolog (e.g., [2, 5, 6]) or use a general constraint system ([16]).

In this paper we provide for parsing with grammars expressed in a general TFS-based formalism, a restriction of ALE ([2]). Our motivation is the design of an abstract (WAM-like) machine for the formalism ([14]); we consider parsing as a computational process and use it as an operational semantics to guide the design of the control structures for the abstract machine. In this paper the machine is not discussed further.

Section 2 outlines the theory of TFSs of [1, 3]. We emphasize **abstract typed feature structures** (AFSs) that encode the essential information of TFSs and extend unification to AFSs. Section 3 introduces an explicit construct of **multi-rooted feature structures** (MRSs) that naturally extend TFSs, used to represent phrasal signs as well as grammar rules. Abstraction is extended to MRSs and the mathematical foundations needed for manipulating them is given. In section 4 a simple bottom-up chart parser for the TFS-based formalism is presented and shown correct. The appendix contains examples of MRSs and grammars as well as a simulation of parsing. Due to space limitations we replace many proofs by informal descriptions and examples; the formal details are given in [15]. The main contributions of this paper are:

- Formalization and explication of the notion of multi-rooted feature structures that are used implicitly in the computational linguistics literature;

- Concise definitions of a TFS-based linguistic formalism, based on abstract MRSs;

- Specification and correctness proofs for parsing in this framework.

## 2 Theory of Feature Structures

### 2.1 Types, Features and Feature Structures

We assume familiarity with the theory of TFS as in [3, chapters 1-6], and only summarize some of its preliminary notions. When dealing with partial functions the symbol '$f(x) \downarrow$' means that $f$ is defined for the value $x$ and the symbol '$\uparrow$' means undefinedness. Whenever the result of an application of a partial function is used as an operand, it is meant that the function is defined for its arguments.

For the following discussion, fix non-empty, finite, disjoint sets TYPES and FEATS of types and feature names, respectively. Let PATHS = FEATS$^*$ denote the collection of **paths**, where FEATS is totally ordered. Fix also an infinite set NODES of nodes and a typing function $\theta$ : NODES $\rightarrow$ TYPES. The set NODES is 'rich' in the sense that for every $t \in$ TYPES, the set $\{q \in$ NODES $\mid \theta(q) = t\}$ is infinite. We use the bounded complete partial order $\sqsubseteq$ over TYPES$\times$TYPES to denote the **type hierarchy**, and the partial function $Approp$ : FEATS $\times$ TYPES $\rightarrow$ TYPES to denote the **appropriate specification**.

A **feature structure** is a directed, connected, labeled graph consisting of a finite, nonempty set of nodes $Q \subseteq$ NODES, a root $\bar{q} \in Q$, and a partial function $\delta : Q \times$ FEATS $\rightarrow Q$ specifying the arcs such that every node $q \in Q$ is accessible from $\bar{q}$. We overload '$\sqsubseteq$' to denote also subsumption of feature structures. Two feature structures $A_1$ and $A_2$ are **alphabetic variants** ($A_1 \sim A_2$) iff $A_1 \sqsubseteq A_2$ and $A_2 \sqsubseteq A_1$.

Alphabetic variants have exactly the same structure, and corresponding nodes have the same types. Only the identities of the nodes distinguish them. The essential properties of a feature structure, excluding the identities of its nodes, can be captured by three components: the set of paths, the type assigned to every path, and the sets of paths that lead to the same node. In contrast to other approaches (e.g., [3]), we first define abstract feature structures and then show their relation to concrete ones. The representation of graphs as sets of paths is inspired by works on the semantics of concurrent programming languages, and the notion of fusion-closure is due to [4].

**Definition 2.1 (Abstract feature structures)** *A pre- abstract feature structure (pre-AFS) is a triple $\langle \Pi, \Theta, \approx \rangle$, where*

- *$\Pi \subseteq$ PATHS is a non-empty set of paths*

- *$\Theta : \Pi \rightarrow$ TYPES is a total function, assigning a type to every path*

- *$\approx \; \subseteq \Pi \times \Pi$ is a relation specifying reentrancy (with $[\approx]$ the set of its equivalence classes)*

*An **abstract feature structure** (AFS) is a pre-AFS for which the following requirements hold:*

- *$\Pi$ is prefix-closed: if $\pi\alpha \in \Pi$ then $\pi \in \Pi$ (where $\pi, \alpha \in$ PATHS)*

- *A is fusion-closed: if $\pi\alpha \in \Pi$ and $\pi'\alpha' \in \Pi$ and $\pi \approx \pi'$ then $\pi\alpha' \in \Pi, \Theta(\pi\alpha') = \Theta(\pi'\alpha')$ (as well as $\pi'\alpha \in \Pi, \Theta(\pi'\alpha) = \Theta(\pi\alpha)$), and $\pi\alpha' \approx \pi'\alpha'$ (as well as $\pi'\alpha \approx \pi\alpha$)*

- *$\approx$ is an equivalence relation with a finite index*

- *$\Theta$ respects the equivalence: if $\pi_1 \approx \pi_2$ then $\Theta(\pi_1) = \Theta(\pi_2)$*

An AFS $\langle \Pi, \Theta, \approx \rangle$ is *well-typed* if $\Theta(\pi) \neq \top$ for every $\pi \in \Pi$ and if $\pi f \in \Pi$ then $Approp(f, \Theta(\pi))\downarrow$ and $Approp(f, \Theta(\pi)) \sqsubseteq \Theta(\pi f)$. It is *totally well typed* if, in addition, for every $\pi \in \Pi$, if $Approp(f, \Theta(\pi))\downarrow$ then $\pi f \in \Pi$.

Abstract features structures can be related to concrete ones in a natural way: If $A = (Q, \bar{q}, \delta)$ is a TFS then $Abs(A) = \langle \Pi_A, \Theta_A, \approx_A \rangle$ is defined by:

- $\Pi_A = \{\pi \mid \delta(\bar{q}, \pi)\downarrow\}$
- $\Theta_A(\pi) = \theta(\delta(\bar{q}, \pi))$
- $\pi_1 \approx_A \pi_2$ iff $\delta(\bar{q}, \pi_1) = \delta(\bar{q}, \pi_2)$

It is easy to see that $Abs(A)$ is an abstract feature structure.

For the reverse direction, consider an AFS $A = \langle \Pi, \Theta, \approx \rangle$. First construct a 'pseudo-TFS', $Conc(A) = (Q, \bar{q}, \delta)$, that differs from a TFS only in that its nodes are not drawn from the set NODES. Let $Q = \{q_{[\pi]} \mid [\pi] \in [\approx]\}$. Let $\theta(q_{[\pi]}) = \Theta(\pi)$ for every node – since $A$ is an AFS, $\Theta$ respects the equivalence and therefore $\theta$ is representative-independent. Let $\bar{q} = q_{[\epsilon]}$ and $\delta(q_{[\pi]}, f) = q_{[\pi f]}$ for every node $q_{[\pi]}$ and feature $f$. Since $A$ is fusion-closed, $\delta$ is representative-independent. By injecting $Q$ into NODES making use of the richness on NODES, a concrete TFS $Conc(A)$ is obtained, representing the equivalence class of alphabetic variants that can be obtained that way. We abuse the notation $Conc(A)$ in the sequel to refer to this set of alphabetic variants.

**Theorem 2.2** *If $A' \in Conc(A)$ then $Abs(A') = A$.*

AFSs can be partially ordered: $\langle \Pi_A, \Theta_A, \approx_A \rangle \preceq \langle \Pi_B, \Theta_B, \approx_B \rangle$ iff $\Pi_A \subseteq \Pi_B, \approx_A \subseteq \approx_B$ and for every $\pi \in \Pi_A, \Theta_A(\pi) \sqsubseteq \Theta_B(\pi)$. This order corresponds to the subsumption ordering on TFSs, as the following theorems show.

**Theorem 2.3** $A \sqsubseteq B$ *iff $Abs(A) \preceq Abs(B)$.*

**Theorem 2.4** *For every $A \in Conc(A'), B \in Conc(B'), A \sqsubseteq B$ iff $A' \preceq B'$.*

**Corollary 2.5** $A \sim B$ *iff $Abs(A) = Abs(B)$.*

**Corollary 2.6** $Conc(A') \sim Conc(B')$ *iff $A = B$.*

## 2.2 Unification

As there exists a one to one correspondence between AFSs and (alphabetic variants of) concrete ones, we define unification over AFSs. This leads to a simpler definition that captures the essence of the operation better than the traditional definition. We use the term 'unification' to refer to both the operation and its result.

**Definition 2.7 (Closure operations)** *Let $Cl$ be a fusion-closure operation on pre-AFSs: $Cl(A) = A'$, where $A'$ is the least extension of $A$ to a fusion-closed structure. Let $Eq(\langle \Pi, \Theta, \approx \rangle) = \langle \Pi, \Theta, \approx' \rangle$ where $\approx'$ is the least extension of $\approx$ to an equivalence relation. Let $Ty(\langle \Pi, \Theta, \approx \rangle) = \langle \Pi, \Theta', \approx \rangle$ where $\Theta'(\pi) = \bigsqcup_{\pi' \approx \pi} \Theta(\pi)$.*

**Definition 2.8 (Unification)** *The unification $A \sqcup B$ of two AFSs $A = \langle \Pi_A, \Theta_A, \approx_A \rangle$ and $B = \langle \Pi_B, \Theta_B, \approx_B \rangle$ is an AFS $C' = Ty(Eq(Cl(C)))$, where:*

- $C = \langle \Pi_C, \Theta_C, \approx_C \rangle$
- $\Pi_C = \Pi_A \cup \Pi_B$
- $\Theta_C(\pi) = \begin{cases} \Theta_A(\pi) \sqcup \Theta_B(\pi) & \text{if } \pi \in \Pi_A \text{ and } \pi \in \Pi_B \\ \Theta_A(\pi) & \text{if } \pi \in \Pi_A \text{ only} \\ \Theta_B(\pi) & \text{if } \pi \in \Pi_B \text{ only} \end{cases}$

- $\approx_C = \approx_A \cup \approx_B$

The unification **fails** if there exists a path $\pi \in \Pi_{C'}$ such that $\Theta_{C'}(\pi) = \top$.

**Lemma 2.9** *Cl preserves prefixes: If $A$ is a prefix-closed pre-AFS and $A' = Cl(A)$ then $A'$ is prefix-closed.*

**Lemma 2.10** *Eq preserves prefixes and fusions: If $A$ is a prefix- and fusion-closed pre-AFS and $A' = Eq(A)$ then $A'$ is prefix- and fusion-closed.*

**Corollary 2.11** *If $A$ and $B$ are AFSs, then so is $A \sqcup B$.*

$C'$ is the smallest AFS that contains $\Pi_C$ and $\approx_C$. Since $\Pi_A$ and $\Pi_B$ are prefix-closed, so is $\Pi_C$. However, $\Pi_C$ and $\approx_C$ might not be fusion-closed. This is why $Cl$ is applied to them. As a result of its application, new paths and equivalence classes might be added. By lemma 2.9, if a path is added all its prefixes are added, too, so the prefix-closure is preserved. Then, $Eq$ extends $\approx$ to an equivalence relation, without harming the prefix- and fusion-closure properties (by lemma 2.10). Finally, $Ty$ sees to it that $\Theta$ respects the equivalences.

**Lemma 2.12** *Unification is commutative: $A \sqcup B = B \sqcup A$.*

**Lemma 2.13** *Unification is associative: $(A \sqcup B) \sqcup C = A \sqcup (B \sqcup C)$.*

The result of a unification can differ from any of its arguments in three ways: paths that were not present can be added; the types of nodes can become more specific; and reentrancies can be added, that is, the number of equivalence classes of paths can decrease. Consequently, the result of a unification is always more specific than any of its arguments.

**Theorem 2.14** *If $C' = A \sqcup B$ then $A \preceq C'$.*

TFSs (and therefore AFSs) can be seen as a generalization of first-order terms (FOTs) (see [1]). Accordingly, AFS unification resembles FOT unification; however, the notion of *substitution* that is central to the definition of FOT unification is missing here, and as far as we know, no analog to substitutions in the domain of feature structures was ever presented.

# 3 Multi-rooted Structures

To be able to represent complex linguistic information, such as phrase structure, the notion of feature structures has to be extended. HPSG does so by introducing special features, such as DTRS (daughters), to encode trees in TFSs. This solution requires a declaration of the special features, along with their intended meaning; such a declaration is missing in [10]. An alternative technique is employed by Shieber ([11]): natural numbers are used as special features, to encode the order of daughters in a tree. In a typed system this method necessitates the addition of special types as well; theoretically, the number of features and types necessary to state rules is unbounded.

As a more coherent, mathematically elegant solution, we define multi-rooted structures, naturally extending TFSs. These structures provide a means to represent phrasal signs and grammar rules. They are used implicitly in the computational linguistics literature, but to the best of our knowledge no explicit, formal theory of these structures and their properties was formulated before.

**Definition 3.1 (Multi-rooted structures)** *A **multi-rooted feature structure** (MRS) is a pair $\langle \bar{Q}, G \rangle$ where $G = \langle Q, \delta \rangle$ is a finite, directed, labeled graph consisting of a set $Q \subseteq \textsc{Nodes}$ of nodes and a partial function $\delta : Q \times \textsc{Feats} \to Q$ specifying the arcs, and where $\bar{Q}$ is an ordered, non-empty (repetition-free) list of distinguished nodes in $Q$ called **roots**. $G$ is not necessarily connected, but the union of all the nodes reachable from all the roots in $\bar{Q}$ is required to yield exactly $Q$. The **length** of a MRS is the number of its roots, $|\bar{Q}|$.*

Meta-variables $\sigma, \rho$ range over MRSs, and $\delta, Q$ and $\bar{Q}$ over their constituents. If $\langle \bar{Q}, G \rangle$ is a MRS and $\bar{q}_i$ is a root in $\bar{Q}$ then $\bar{q}_i$ naturally induces a feature structure $Pr(\bar{Q}, i) = (Q_i, \bar{q}_i, \delta_i)$, where $Q_i$ is the set of nodes reachable from $\bar{q}_i$ and $\delta_i = \delta|_{Q_i}$.

One can view a MRS $\langle \bar{Q}, G \rangle$ as an ordered sequence $\langle A_1, \ldots, A_n \rangle$ of (not necessarily disjoint) feature structures, where $A_i = Pr(\bar{Q}, i)$ for $1 \leq i \leq n$. Note that such an ordered list of feature structures is not a sequence in the mathematical sense: removing an element from the list effects the other elements (due to reentrancy among elements). Nevertheless, we can think of a MRS as a sequence where a subsequence is obtained by taking a subsequence of the roots and considering only the feature structures they induce. We use the two views interchangeably. Figure 1 depicts a MRS and its view as a sequence of feature structures.

A MRS is well-typed if all its constituent feature structures are well-typed, and is totally well-typed if all its constituents are. Subsumption is extended to MRSs as follows:

**Definition 3.2 (Subsumption of multi-rooted structures)** *A MRS $\sigma = \langle \bar{Q}, G \rangle$ subsumes a MRS $\sigma' = \langle \bar{Q}', G' \rangle$ (denoted by $\sigma \sqsubseteq \sigma'$) if $|\bar{Q}| = |\bar{Q}'|$ and there exists a total function $h : Q \to Q'$ such that:*

- *for every root $\bar{q}_i \in \bar{Q}, h(\bar{q}_i) = \bar{q}_i'$*
- *for every $q \in Q$, $\theta(q) \sqsubseteq \theta'(h(q))$*
- *for every $q \in Q$ and $f \in \text{Feats}$, if $\delta(q, f)\downarrow$ then $h(\delta(q, f)) = \delta'(h(q), f)$*

We define abstract multi-rooted structures in an analog way to abstract feature structures.

**Definition 3.3 (Abstract multi-rooted structures)** *A pre- abstract multi rooted structure (pre-AMRS) is a quadruple $A = \langle Ind, \Pi, \Theta, \approx \rangle$, where:*

- *$Ind$, the **indices** of $A$, is the sequence $\langle 1, \ldots, n \rangle$ for some $n$*
- *$\Pi \subseteq Ind \times \text{Paths}$ is a non-empty set of indexed paths, such that for each $i \in Ind$ there exists some $\pi \in \text{Paths}$ that $(i, \pi) \in \Pi$.*
- *$\Theta : \Pi \to \text{Types}$ is a total type-assignment function*
- *$\approx \subseteq \Pi \times \Pi$ is a relation*

*An **abstract multi-rooted structure** (AMRS) is a pre-AMRS $A$ for which the following requirements, naturally extending those of AFSs, hold:*

- *$\Pi$ is prefix-closed: if $(i, \pi\alpha) \in \Pi$ then $(i, \pi) \in \Pi$*
- *$A$ is fusion-closed: if $(i, \pi\alpha) \in \Pi$ and $(i', \pi'\alpha') \in \Pi$ and $(i, \pi) \approx (i', \pi')$ then $(i, \pi\alpha') \in \Pi$ (as well as $(i', \pi'\alpha) \in \Pi$), and $(i, \pi\alpha') \approx (i', \pi'\alpha')$ (as well as $(i', \pi'\alpha) \approx (i, \pi\alpha)$)*
- *$\approx$ is an equivalence relation*
- *$\Theta$ respects the equivalence: if $(i_1, \pi_1) \approx (i_2, \pi_2)$ then $\Theta(i_1, \pi_1) = \Theta(i_2, \pi_2)$*

An AMRS $\langle Ind, \Pi, \Theta, \approx \rangle$ is well-typed if for every $(i, \pi) \in \Pi$, $\Theta(i, \pi) \neq \top$ and if $(i, \pi f) \in \Pi$ then $Approp(f, \Theta(i, \pi))\downarrow$ and $Approp(f, \Theta(i, \pi)) \sqsubseteq \Theta(i, \pi f)$. It is totally well typed if, in addition, for every $(i, \pi) \in \Pi$, if $Approp(f, \Theta(i, \pi))\downarrow$ then $(i, \pi f) \in \Pi$. The **length** of an AMRS $A$ is $len(A) = |Ind_A|$.

The closure operations $Cl$ and $Eq$ are naturally extended to AMRSs: If $A$ is a pre-AMRS then $Cl(A)$ is the least extension of $A$ that is prefix- and fusion-closed, and $Eq(A)$ is the least extension of $A$ to a pre-AMRS in which $\approx$ is an equivalence relation. In addition, $Ty(\langle Ind, \Pi, \Theta, \approx \rangle) = \langle Ind, \Pi, \Theta', \approx \rangle$ where $\Theta'(i, \pi) = \bigsqcup_{(i', \pi') \approx (i, \pi)} \Theta(i', \pi')$. The partial order $\preceq$ is extended to AMRSs: $\langle Ind_A, \Pi_A, \Theta_A, \approx_A \rangle \preceq \langle Ind_B, \Pi_B, \Theta_B, \approx_B \rangle$ iff $Ind_A = Ind_B, \Pi_A \subseteq \Pi_B, \approx_A \subseteq \approx_B$ and for every $(i, \pi) \in \Pi_A, \Theta_A(i, \pi) \sqsubseteq \Theta_B(i, \pi)$.

AMRSs, too, can be related to concrete ones in a natural way: If $\sigma = \langle \bar{Q}, G \rangle$ is a MRS then $Abs(\sigma) = \langle Ind_\sigma, \Pi_\sigma, \Theta_\sigma, \approx_\sigma \rangle$ is defined by:

- $Ind_\sigma = \langle 1, \ldots, |\bar{Q}| \rangle$
- $\Pi_\sigma = \{(i, \pi) \mid \delta(\bar{q}_i, \pi)\downarrow\}$
- $\Theta_\sigma(i, \pi) = \theta(\delta(\bar{q}_i, \pi))$
- $(i, \pi_1) \approx_\sigma (j, \pi_2)$ iff $\delta(\bar{q}_i, \pi_1) = \delta(\bar{q}_j, \pi_2)$

It is easy to see that $Abs(\sigma)$ is an AMRS. In particular, notice that for every $i \in Ind_\sigma$ there exists a path $\pi$ such that $(i, \pi) \in \Pi_\sigma$ since for every $i, \delta(\bar{q}_i, \epsilon)\downarrow$. The reverse operation, $Conc$, can be defined in a similar manner.

AMRSs are used to represent ordered collections of AFSs. However, due to the possibility of value sharing among the constituents of AMRSs, they are not sequences in the mathematical sense, and the notion of sub-structure has to be defined in order to relate them to AFSs.

**Definition 3.4 (Sub-structures)** *Let $A = \langle Ind_A, \Pi_A, \Theta_A, \approx_A \rangle$; let $Ind_B$ be a finite (contiguous) subsequence of $Ind_A$; let $n+1$ be the index of the first element of $Ind_B$. The **sub-structure** of $A$ induced by $Ind_B$ is an AMRS $B = \langle Ind_B, \Pi_B, \Theta_B, \approx_B \rangle$ such that:*

- $(i - n, \pi) \in \Pi_B$ iff $i \in Ind_B$ and $(i, \pi) \in A$
- $\Theta_B(i - n, \pi) = \Theta_A(i, \pi)$ if $i \in Ind_B$
- $(i_1 - n, \pi_1) \approx_B (i_2 - n, \pi_2)$ iff $i_1 \in Ind_B, i_2 \in Ind_B$ and $(i_1, \pi_1) \approx_A (i_2, \pi_2)$

A sub-structure of $A$ is obtained by selecting a subsequence of the indices of $A$ and considering the structure they induce. Trivially, this structure is an AMRS. We use $A^{j..k}$ to refer to the sub-structure of $A$ induced by $\langle j, \ldots, k \rangle$. If $Ind_B = \{i\}$, $A^{i..i}$ can be identified with an AFS, denoted $A^i$.

The notion of concatenation has to be defined for AMRSs, too:

**Definition 3.5 (Concatenation)** *The **concatenation** of $A = \langle Ind_A, \Pi_A, \Theta_A, \approx_A \rangle$ and $B = \langle Ind_B, \Pi_B, \Theta_B, \approx_B \rangle$ of lengths $n_A, n_B$, respectively (denoted by $A \cdot B$), is an AMRS $C = \langle Ind_C, \Pi_C, \Theta_C, \approx_C \rangle$ such that*

- $Ind_C = \langle 1, \ldots, n_A + n_B \rangle$
- $\Pi_C = \Pi_A \cup \{(i + n_A, \pi) \mid (i, \pi) \in \Pi_B\}$
- $\Theta_C(i, \pi) = \begin{cases} \Theta_A(i, \pi) & \text{if } i \leq n_A \\ \Theta_B(i - n_A, \pi) & \text{if } i > n_A \end{cases}$
- $\approx_C = \approx_A \cup \{((i_1 + n_A, \pi_1), (i_2 + n_A, \pi_2)) \mid (i_1, \pi_1) \approx_B (i_2, \pi_2)\}$

We now extend the definition of unification to AMRSs: we want to allow the unification of two $AFSs$, one of which is a part of an AMRS. Therefore, one operand is a pair consisting of an AMRS and an index, specifying some element of it, and the second operand is an AFS. Recall that due to reentrancies, other elements of the AMRS can be affected by this operation. Therefore, the result of the unification is a new AMRS.

**Definition 3.6 (Unification in context)** *Let $A = \langle Ind_A, \Pi_A, \Theta_A, \approx_A \rangle$ be an AMRS, $B = \langle \Pi_B, \Theta_B, \approx_B \rangle$ an AFS. $(A, j) \sqcup B$ is defined if $j \in Ind_A$, in which case it is the AMRS $C' = Ty(Eq(Cl(\langle Ind_C, \Pi_C, \Theta_C, \approx_C \rangle)))$, where*

- $Ind_C = Ind_A$
- $\Pi_C = \Pi_A \cup \{(j, \pi) \mid \pi \in \Pi_B\}$
- $\Theta_C(i, \pi) = \begin{cases} \Theta_A(i, \pi) & \text{if } i \neq j \\ \Theta_A(i, \pi) \sqcup \Theta_B(\pi) & \text{if } i = j \text{ and } (i, \pi) \in \Pi_A \text{ and } \pi \in \Pi_B \\ \Theta_A(i, \pi) & \text{if } i = j \text{ and } (i, \pi) \in \Pi_A \text{ and } \pi \notin \Pi_B \\ \Theta_B(\pi) & \text{if } i = j \text{ and } (i, \pi) \notin \Pi_A \text{ and } \pi \in \Pi_B \end{cases}$

- $\approx_C \, = \, \approx_A \cup \{((j, \pi_1), (j, \pi_2)) \mid \pi_1 \approx_B \pi_2\}$

*The unification fails if there exists some pair $(i, \pi) \in \Pi_{C'}$ such that $\Theta_{C'}(i, \pi) = \top$.*

Many of the properties of AFSs, proven in the previous section, hold for AMRSs, too. In particular, if $A$ is an AMRSs then so is $(A, j) \sqcup B$ if it is defined, $len((A, j) \sqcup B) = len(A)$ and $(A, j) \sqcup B \sqsupseteq A$.

## 4 Parsing

Parsing is the process of determining whether a given string belongs to the language defined by a given grammar, and assigning a structure to the permissible strings. We formalize and explicate some of the notions of [3, chapter 13]. We give direct definitions for rules, grammars and languages, based on our new notion of AMRSs. This presentation is more adequate to current TFS-based systems than [7, 12], that use first-order terms. Moreover, it does not necessitate special, ad-hoc features and types for encoding trees in TFSs as [11] does. We don't assume any explicit context-free back-bone for the grammars, as does [13].

We describe a pure bottom-up chart-based algorithm. The formalism we presented is aimed at being a platform for specifying grammars in HPSG, which is characterized by employing a few very general rules; selecting the rules that are applicable in every step of the process requires unification anyhow. Therefore we choose a particular parsing algorithm that does not make use of top down predictions but rather assumes that every rule might be applied in every step. This assumption is realized by initializing the chart with predictive edges for every rule, in every position.

### 4.1 Rules and Grammars

We define rules and grammars over a fixed set WORDS of words. However, we assume that the lexicon associates with every word $w$ a feature structure $C(w)$, its **category**,[1] so we can ignore the terminal words and consider only their categories. The input for the parser, therefore, is a sequence[2] of TFSs rather than a string of words.

**Definition 4.1 (Pre-terminals)** *Let $w = w_1 \ldots w_n \in \text{WORDS}^*$ and $A_i = C(w_i)$ for $1 \leq i \leq n$. $PT_w(j, k)$ is defined if $1 \leq j \leq k \leq n$, in which case it is the AMRS $Abs(\langle A_j, A_{j+1}, \ldots, A_k \rangle)$. Note that $PT_w(j, k) \cdot PT_w(k+1, m) = PT_w(j, m)$. We omit the subscript $w$ when it is clear from the context.*

**Definition 4.2 (Rules)** *A **rule** is a MRS of length $n > 1$ with a distinguished last element. If $\langle A_1, \ldots, A_{n-1}, A_n \rangle$ is a rule then $A_n$ is its **head**[3] and $\langle A_1, \ldots, A_{n-1} \rangle$ is its **body**.[4] We write such a rule as $\langle A_1, \ldots, A_{n-1} \Rightarrow A_n \rangle$. In addition, every category of a lexical item is a rule (with an empty body). We assume that such categories don't head any other rule.*

**Definition 4.3 (Grammars)** *A **grammar** $G = (\mathcal{R}, A_s)$ is a finite set of rules $\mathcal{R}$ and a **start symbol** $A_s$ that is a TFS.*

An example grammar, whose purpose is purely illustrative, is depicted in figure 2. For the following discussion we fix a particular grammar $G = (\mathcal{R}, A_s)$.

**Definition 4.4 (Derivation)** *An AMRS $A = \langle Ind_A, \Pi_A, \Theta_A, \approx_A \rangle$ **derives** an AMRS $B$ (denoted $A \to B$) if there exists a rule $\rho \in \mathcal{R}$ with $len(\rho) = n$ and $R = Abs(\rho)$, such that*

- *some element of $A$ unifies with the head of $R$: there exist AMRSs $A', R'$ and $j \in Ind_A$ such that $A' = (A, j) \sqcup R^n$ and $R' = (R, n) \sqcup A^j$*

---

[1] Ambiguous words are associated with more than one category. We ignore such cases in the sequel.
[2] We assume that there is no reentrancy among lexical items.
[3] This use of *head* must not be confused with the linguistic one, the core features of a phrase.
[4] Notice that the traditional direction is reversed and that the head and the body need not be disjoint.

- $B$ can be obtained by replacing the $j$-th element of $A'$ with the body of $R'$.[5]

'$\xrightarrow{*}$' is the reflexive transitive closure of '$\rightarrow$'.

Intuitively, $A$ derives $B$ through some AFS $A^j$ in $A$, if some rule $\rho \in \mathcal{R}$ licenses the derivation. $A^j$ is unified with the head of the rule, and if the unification succeeds, the (possibly modified) body of the rule replaces $A^j$ in $A$.

**Definition 4.5 (Language)** *The **language** of a grammar $G$ is $L(G) = \{w = w_1 \cdots w_n \in \text{WORDS}^* \mid A \xrightarrow{*} B \text{ for some } A \sqsupseteq Abs(A_s) \text{ and } B \sqsupseteq PT_w(1,n)\}$.*

Figure 3 shows a sequence of derivations, starting from some feature structure that is more specific than the initial symbol and ending in a sequence of structures that can stand for the string "John loves fish", based upon the example grammar.

## 4.2 Parsing as Operational Semantics

We view parsing as a computational process endowing TFS formalisms with an operational semantics, which can be used to derive control mechanisms for an abstract machine we design ([14]). A computation is triggered by some input string of words $w = w_1 \cdots w_n$ of length $n > 0$. For the following discussion we fix a particular input string $w$ of length $n$. A *state* of the computation is a set of *items*, and states are related by a transition relation. The presentation below corresponds to a pure bottom-up algorithm.

**Definition 4.6 (Dotted rules)** *A **dotted rule** (or **edge**) is a pair $\langle A, k \rangle$ where $A = Abs(\sigma)$ is an AMRS such that $\rho \sqsubseteq \sigma$ for some $\rho \in \mathcal{R}$ and where $0 \le k < len(A)$. An edge $\langle A, k \rangle$ is **complete** if $k = len(A) - 1$; an edge is **active** otherwise.*

A dotted rule consists of an AMRS $A$ that is more specific than (the abstraction of) some grammar rule, and a number $k$ that denotes the position of the **dot** within $A$. The dot can precede any element of $A$ (in the case of lexical rules, it can also succeed the rule).

**Definition 4.7 (Items)** *An **item** is a triple $[i, \langle A, k \rangle, j]$ where $0 \le i \le j \le n$ and $\langle A, k \rangle$ is a dotted rule. An item is **complete** if the edge in it is complete. Let ITEMS be the collection of all items.*

During parsing, the intuitive meaning of an item is that the part of $A$ prior to the dot (which is indicated by $k$) derives a substring of the input, and if it can be shown that the part of $A$ succeeding the dot derives a consecutive substring of the input, then the head of $A$ derives the concatenation of the two substrings. This invariant is formally defined and proven in the section 4.3. $i$ and $j$ indicate the *span* of the item.

A computation is determined by a sequence of states, each of which is a collection of items, where the first state corresponds to the initialization and each subsequent state contains its predecessor and is related to it by the transition relation.

**Definition 4.8 (States)** *A **state** $S \subseteq$ ITEMS is a finite set of items.*

**Definition 4.9 (Initialization)** *Let $\hat{S} = I_{lex} \cup I_{predict}$ be the **initial state**, where:*

$$I_{lex} = \{[i-1, \langle A_i, 0 \rangle, i] \mid 1 \le i \le n \text{ and } A_i = PT_w(i,i)\}$$

$$I_{predict} = \{[i, \langle Abs(\rho), 0 \rangle, i] \mid 0 \le i \le n \text{ and } \rho \in \mathcal{R}\}$$

$I_{lex}$ contains the (complete) items that correspond to categories of the input words, whereas $I_{predict}$ contains an (active) item for each grammar rule and a position in the input string.

---
[5]The exact details can be found in [15].

**Definition 4.10 (Dot movement)** *The partial function* $DM$ : ITEMS × ITEMS → ITEMS *is defined as follows:* $DM([i, \langle A, k_A \rangle, l_A], [l_B, \langle B, k_B \rangle, j]) = [i, \langle C, k_C \rangle, j]$, *where:*

- $l_A = l_B, n = len(A), m = len(B)$
- $k_A < n - 1$ *(the edge* $\langle A, k_A \rangle$ *is active)*, $k_B = m - 1$ *(the edge* $\langle B, k_B \rangle$ *is complete)*, $k_C = k_A + 1$
- $C = (A, k+1) \sqcup B^m$ *(C is obtained from A by unifying the element of A succeeding the dot with the head of B)*

$DM$ is not defined if $l_A \neq l_B$, if the edge in its first argument is complete, if the edge in its second argument is active, or if the unification fails.

**Lemma 4.11** *If* $x = [i_x, \langle A_x, k_x \rangle, j_x] \in$ ITEMS, $y = [i_y, \langle A_y, k_y \rangle, j_y] \in$ ITEMS *and* $z = DM(x, y)$ *is defined, then* $z = [i_z, \langle A_z, k_z \rangle, j_z]$ *where* $i_z = i_x, j_z \geq i_z, A_z \succeq A_x$ *and* $k_z > k_x$.

**Corollary 4.12** *If* $x, y \in$ ITEMS *then* $DM(x, y) \in$ ITEMS *if it is defined.*

To compute the next state, new items are added if they result by applying $DM$ to existing items, unless the result is more specific than existing items. This is a realization of the *subsumption check* suggested in [11, 12].

**Definition 4.13 (Ordering items)** *If* $x = [i_x, \langle A_x, k_x \rangle, j_x]$ *and* $y = [i_y, \langle A_y, k_y \rangle, j_y]$ *are items,* $x$ ***subsumes*** $y$ *(written* $x \preceq y$*) iff* $i_x = i_y, j_x = j_y, k_x = k_y$ *and* $A_x \preceq A_y$.

**Definition 4.14** *Let* $\Delta(S) = \{z \mid z = DM(x, y)$ *for some* $x, y \in S$ *and there does not exist* $z' \in S$ *such that* $z' \preceq z\}$. *The transition relation '*⊢*' holds between two states* $S$ *and* $S'$ *(denoted by* $S \vdash S'$*) if* $S' = S \cup \Delta(S)$.

**Definition 4.15 (Computation)** *A **computation** is an infinite sequence of states* $S_i, i \geq 0$, *such that* $S_0 = \hat{S}$ *and for every* $i \geq 0$, $S_i \vdash S_{i+1}$. *A computation is **terminating** if there exists some* $m \geq 0$ *for which* $S_m = S_{m+1}$ *(i.e., a fix-point is reached). A computation is **successful** if one of its states contains an item of the form* $[0, \langle A, k-1 \rangle, n]$ *where* $n$ *is the input length,* $k = len(A)$ *and* $Abs(A_s) \sqsubseteq A^k$; *otherwise, the computation fails.*

### 4.3 Proof of Correctness

In this section we show that parsing, as defined above, is (partially) correct. First, the algorithm is *sound*: computations succeed only for input strings that are sentences of the language. Second, it is *complete*: if a string is a sentence, it is accepted by the algorithm. Then we show that the computation terminates for *off-line parsable* grammars.

#### 4.3.1 Soundness

**Lemma 4.16** *If* $[i, \langle A, k \rangle, j] \in S_m$ *for some* $m \geq 0$ *then* $i = j$ *only if* $\langle A, k \rangle$ *is not complete and* $k = 0$.

**Proof:** By induction on $m$.

**Theorem 4.17 (Completion)** *If* $\langle A, k \rangle$ *is a complete edge,* $len(A) > 1$ *and* $A^{1..k} \stackrel{*}{\to} B$ *then* $A^{k+1} \stackrel{*}{\to} B$.

**Proof:** Since $\langle A, k \rangle$ is an edge such that $len(A) > 1$, there exists an abstract rule $R$ such that $R \sqsubseteq A$. Hence $(A, k+1) \sqcup R^{k+1} = A^{k+1}$, $(R, k+1) \sqcup A^{k+1} = A$ and $A^{k+1} \to A^{1..k}$. Since $A^{1..k} \stackrel{*}{\to} B$ we obtain $A^{k+1} \stackrel{*}{\to} B$.

**Theorem 4.18 (Parsing invariant (a))** *If* $z = [i_z, \langle A_z, k_z \rangle, j_z] \in S_m$ *for some* $m \geq 0$, $l = len(A_z)$ *and* $i_z < j_z$, *then* $A_z^{1..k_z} \stackrel{*}{\to} PT(i_z + 1, j_z)$ *if* $l > 1$, $A_z^1 \stackrel{*}{\to} PT(i_z + 1, j_z)$ *if* $l = 1$.

**Proof:** By induction on $m$. Base: for all items $z \in I_{predict}$, $i_z = j_z$ and the proposition obtains (vacuously). For all items $z \in I_{lex}$, $l = 1$ and $A_z = PT(i_z + 1, j_z)$.

If $m > 0$: Let $x = DM(x, y)$ where $x = [i_x, \langle A_x, k_x \rangle, j_x]$, $y = [i_y, \langle A_y, k_y \rangle, j_y]$ and $x, y \in \text{ITEMS}$. $\langle A_y, k_y \rangle$ is complete, hence by 4.16 $i_y < j_y$ and by the induction hypothesis and the completion theorem, $A_y^{k_y} \stackrel{*}{\to} PT(i_y + 1, j_y)$. Also, $len(A_z) > 1$ since all rules are of length $> 1$. By the induction hypothesis, $A_x^{1..k_x} \stackrel{*}{\to} PT(i_x + 1, j_x)$. Therefore $A_z^{1..k_x} \stackrel{*}{\to} PT(i_x + 1, j_x)$. Since $A_y^{k_y} \stackrel{*}{\to} PT(i_y + 1, j_y)$, we get $A_z^{k_x+1} \stackrel{*}{\to} PT(i_y + 1, j_y)$. Therefore $A_z^{1..k_z} \stackrel{*}{\to} PT(i_z + 1, j_z)$.

**Corollary 4.19** *If a computation triggered by $w = w_1, \ldots, w_n$ is successful then $w \in L(G)$.*

**Proof:** For a computation to be successful there must be a state $S_m$ that contains some item $[0, \langle A, k-1 \rangle, n]$ where $k = len(A)$ and $Abs(A_s) \sqsubseteq A^k$. From the above theorem it follows that $A^{1..k-1} \stackrel{*}{\to} PT(1, n)$. Since $A$ is complete, by the completion theorem $A^k \stackrel{*}{\to} PT(1, n)$, and thus $w_1, \ldots, w_n \in L(G)$.

### 4.3.2 Completeness

The following lemma shows that one derivation step, licensed by a rule $R$ of length $r + 1$, corresponds to $r$ applications of the $DM$ function, starting with an item that predicts $R$ and advancing the dot $r$ times, until a complete item is generated.

**Lemma 4.20** *If $A \to B$ and there exists $m \geq 0$ such that for every $1 \leq b \leq len(B)$ there exists $x_b \in S_m$ such that $x_b = [i_b, \langle B_b, k_b \rangle, j_b]$ is complete and $B_b^{k_b+1} = B^b$, and $i_1 = 0, j_{b-1} = i_b$ for $0 < b < len(B)$ and $j_{len(B)} = n$, then there exists $m' \geq 0$ such that for every $1 \leq a \leq len(A)$ there exists $y_a \in S_{m'}$ such that $y_a = [i_a, \langle A_a, k_a \rangle, j_a]$ is complete and $A_a^{k_A+1} = A^a$ and $i_1 = 0, j_{a-1} = i_a$ for $0 < a < len(B)$ and $j_{len(B)} = n$.*

**Proof:** (sketch) $A \to B$ by some rule $R$ of length $r+1$ that expands the $p$-th element of $A$ to the elements $q_1, \ldots, q_r$ in $B$. For all elements of $A$ except $p$, the proposition holds by assumption. Since $R$ is a rule, there exists an item $x_R \in I_{predict}$ that $x_R = [i_{q_1}, \langle R, 0 \rangle, i_{q_1}]$ where $i_{q_1}$ is the first index of $x_{q_1}$. Let $y_1 = DM(x_R, x_{q_1})$ and $y_l = DM(y_{l-1}, x_{q_l})$ for $1 < l \leq r$. All the $y$ items exist: by the requirements on the indices of the $x_b$-s, the indices of the $y$ items fit. The unifications performed by $DM$ don't fail: if they would, $A$ wouldn't derive $B$. Then $y_r \in S_{m+r}$ is complete (as there were exactly $r$ applications of $DM$) and $y_r^{r+1} = A^p$.

**Theorem 4.21 (Parsing invariant (b))** *If $A \stackrel{*}{\to} PT(i+1, j)$ for $i < j$ then there exist $m \geq 0$ and $x \in S_m$ such that $x = [i, \langle B, k-1 \rangle, j]$ where $k = len(B)$ and $B^k = A$.*

**Proof:** By induction on $d$, the number of derivation steps. Base: if $d = 0$ then $A \to PT(i+1, j)$ iff $PT(i+1, j) \in I_{lex}$, in which case an item as required exists. If $d > 1$, an immediate application of the above lemma to the induction hypothesis gives the required result.

**Corollary 4.22** *If $w = w_1, \ldots, w_n \in L(G)$ then the computation triggered by $w$ is successful.*

**Proof:** Since $w_1, \ldots, w_n \in L(G)$, there exists an AFS $A \sqsupseteq Abs(A_s)$ such that $A \stackrel{*}{\to} PT(1, n)$. By the parsing invariant, there exist $m \geq 0$ and $x \in S_m$, $x = [0, \langle B, k-1 \rangle, n]$ where $k = len(B)$ and $A = B^k$. $Abs(A_s) \sqsubseteq A = B^k$ and therefore the computation is successful.

### 4.3.3 Termination

It is well-known (see, e.g.,[9, 11]) that unification-based grammar formalisms are Turing-equivalent, and therefore decidability cannot be guaranteed in the general case. This is also true for the formalism we describe here. However, for grammars that satisfy a certain restriction, termination of the computation can be proven. The following definition is an adaptation of the one given in [11].

**Definition 4.23 (Off-line parsability)** *A grammar $G$ is **off-line parsable** if there exists a function $f$ from AMRSs to AMRSs such that:*

- *for every AMRS $A$, $f(A) \sqsubseteq A$*

- *the range of $f$ is finite*

- *for every string $w$ and AMRSs $A, B$, if there exist $k_A, k_B$, that $A^{1..k_A} \stackrel{*}{\rightharpoonup} PT_w(i,j)$ and $B^{1..k_B} \stackrel{*}{\rightharpoonup} PT_w(i,j)$ and $A \not\sqsubseteq B$ and $B \not\sqsubseteq A$ then $f(A) \neq f(B)$.*

**Theorem 4.24** *If $G$ is off-line parsable then every computation terminates.*

**Proof:**(sketch) Select some computation triggered by some input $w$ of length $n$. We want to show that only a finite number of items can be generated during the computation. Observe that the indices that determine the span of the item are limited: $0 \leq i \leq j \leq n$. The location of the dot within each AMRS $A$ is also limited: $0 \leq k < len(A)$. It remains to show that only a finite number of edges are generated. Suppose that $[i, \langle A, k \rangle, j] \in S$ is an item that was generated during the computation. Now suppose another item is generated where only the AMRS is different: $[i, \langle B, k \rangle, j]$. If $B \sqsupseteq A$ it will not be included in $\Delta(S)$ because of the subsumption test. There is only a finite number of AMRSs $A'$ such that $B \sqsubseteq A$ (since subsumption is a well-founded relation). Now suppose $A \not\sqsubseteq B$ and $B \not\sqsubseteq A$. By the parsing invariant (a) there exist $A', B'$ such that $A^{1..k} \stackrel{*}{\rightharpoonup} PT_w(i,j)$ and $B^{1..k} \stackrel{*}{\rightharpoonup} PT_w(i,j)$. Since $G$ is off-line parsable, $f(A) \neq f(B)$. Since the range of $f$ is finite, there are only finitely many edges with equal span that are pairwise incomparable.

Since only a finite number of items can be generated, and the states of the computation are such that $S_m \subseteq S_{m+1}$ for $m \geq 0$, a fix-point is reached within a finite number of state transitions.

# A Examples

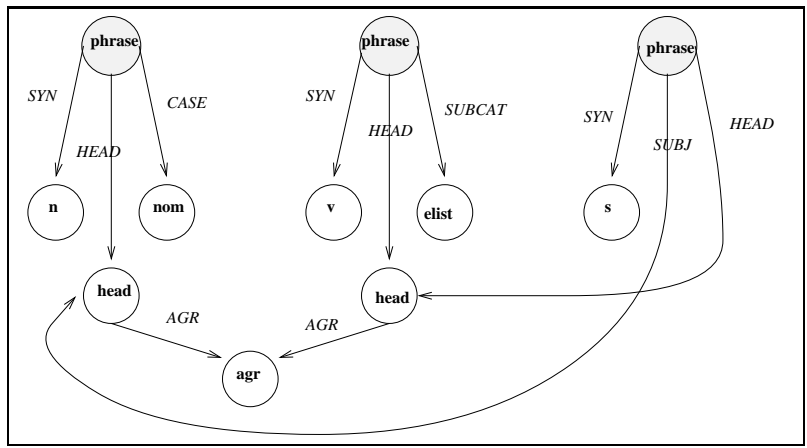

Figure 1: A graph- and AVM- representation of a MRS

The grammar listed in figure 2 consists of four rules and three lexical entries. The rules are extensions of the common context-free rules S → NP VP, NP → PN and VP → V NP. Notice that the head of each rule is on the right hand side of the '⇒' sign. Note also that values are shared among the body and the head of each rule, thus enabling percolation of information during derivation.

Initial symbol:
$$\begin{bmatrix} \text{phrase} \\ SYN: & [\text{s}] \end{bmatrix}$$

Rules:

$$\begin{bmatrix} \text{phrase} \\ SYN: & [\text{n}] \\ HEAD: & \boxed{1} \begin{bmatrix} \text{head} \\ AGR: & \boxed{3} \end{bmatrix} \\ CASE: & [\text{nom}] \end{bmatrix} \begin{bmatrix} \text{phrase} \\ SYN: & [\text{v}] \\ HEAD: & \boxed{2} \begin{bmatrix} \text{head} \\ AGR: & \boxed{3} \end{bmatrix} \\ SBCT: & [\text{elist}] \end{bmatrix} \Rightarrow \begin{bmatrix} \text{phrase} \\ SYN: & [\text{s}] \\ SUBJ: & \boxed{1} \\ HEAD: & \boxed{2} \end{bmatrix} \quad (1)$$

$$\begin{bmatrix} \text{word} \\ SYN: & [\text{pn}] \\ HEAD: & \boxed{1} \\ CASE: & \boxed{2} \end{bmatrix} \Rightarrow \begin{bmatrix} \text{phrase} \\ SYN: & [\text{n}] \\ HEAD: & \boxed{1} \\ CASE: & \boxed{2} \end{bmatrix} \quad (2)$$

$$\begin{bmatrix} \text{word} \\ SYN: & [\text{v}] \\ HEAD: & \boxed{1} \\ SBCT: & \begin{bmatrix} \text{nelist} \\ 1ST: & \boxed{3} \\ RST: & \boxed{2} \end{bmatrix} \end{bmatrix} \boxed{3} \begin{bmatrix} \text{phrase} \\ SYN: & [\text{n}] \\ HEAD: & [\text{head}] \\ CASE: & [\text{acc}] \end{bmatrix} \Rightarrow \begin{bmatrix} \text{phrase} \\ SYN: & [\text{v}] \\ HEAD: & \boxed{1} \\ SBCT: & \boxed{2} \end{bmatrix} \quad (3)$$

Lexicon:

$$``John" \mapsto \begin{bmatrix} \text{word} \\ SYN: & [\text{pn}] \\ HEAD: & \begin{bmatrix} \text{head} \\ AGR: & \begin{bmatrix} \text{agr} \\ PERS: & [\text{3rd}] \\ NUM: & [\text{sg}] \end{bmatrix} \end{bmatrix} \\ CASE: & [\text{case}] \end{bmatrix} \quad (4)$$

$$``loves" \mapsto \begin{bmatrix} \text{word} \\ SYN: & [\text{v}] \\ HEAD: & \begin{bmatrix} \text{head} \\ AGR: & \begin{bmatrix} \text{agr} \\ NUM: & [\text{sg}] \end{bmatrix} \end{bmatrix} \\ SBCT: & \begin{bmatrix} \text{nelist} \\ 1ST: & \begin{bmatrix} \text{phrase} \\ SYN: & [\text{n}] \end{bmatrix} \\ RST: & [\text{elist}] \end{bmatrix} \end{bmatrix} \quad (5)$$

$$``fish" \mapsto \begin{bmatrix} \text{word} \\ SYN: & [\text{pn}] \\ HEAD: & \begin{bmatrix} \text{head} \\ AGR: & [\text{agr}] \end{bmatrix} \\ CASE: & [\text{case}] \end{bmatrix} \quad (6)$$

Figure 2: An example grammar

A leftmost derivation of the string "John loves fish" is given in figure 3, where each derivation bears the number of the rule that licenses it. The string is a sentence of the grammar since the derivation starts with a feature structure that is more specific than the initial symbol and ends with feature structures that are subsumed by the lexical entries of the input string.

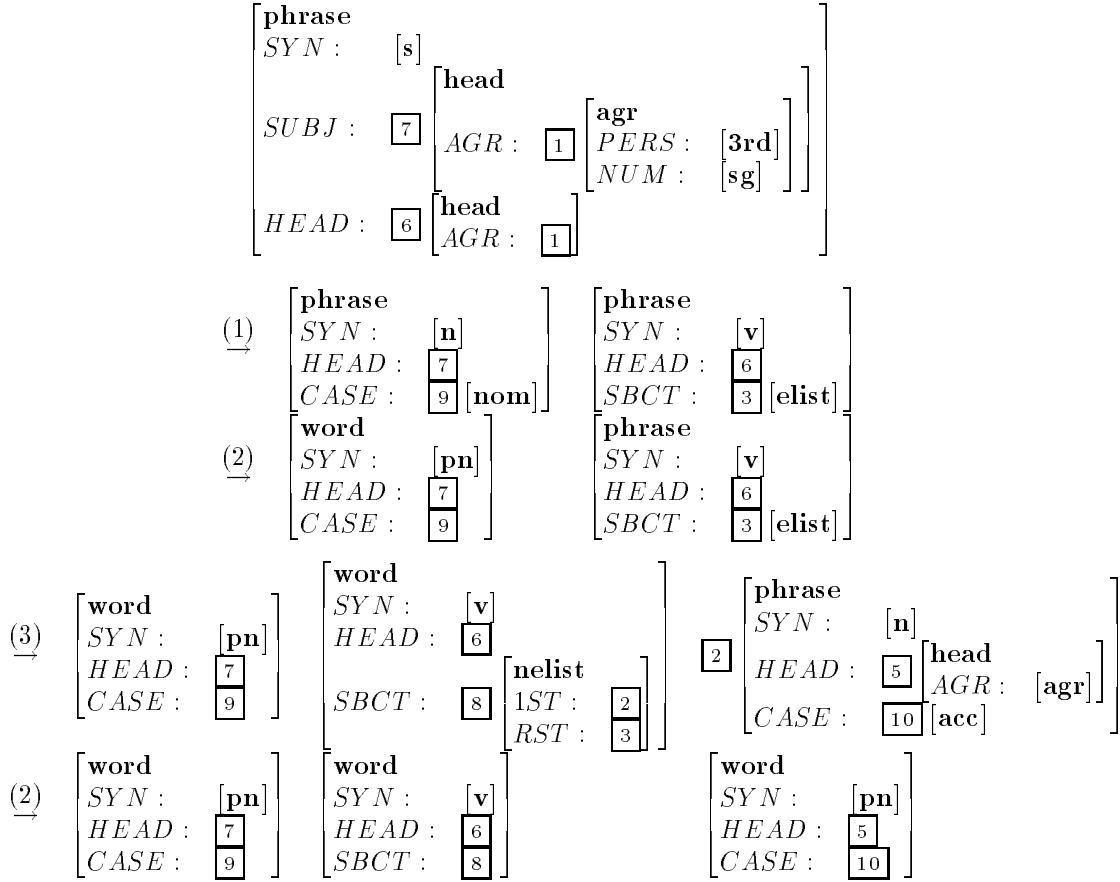

Figure 3: A leftmost derivation

Finally, we simulate the process of parsing with the example grammar and the input "John loves fish". As a matter of convention, if $[i, \langle A, k \rangle, j]$ is an item, we say that the edge $\langle A, k \rangle$ is in the $(i, j)$ entry. We also explicitly indicate the position of the dot (denoted by '•') within MRSs instead of using an integer.

The first state, $S_0$, consists of $I_{lex} \cup I_{predict}$; $I_{lex}$ contains three items: the pre-terminals corresponding to "John", "loves" and "fish" with the dot set to 0 in entries $(0,1), (1,2)$ and $(2,3)$, respectively. $I_{predict}$ contains, for each entry $(i,i)$, where $0 \leq i \leq 3$, an edge of the form $\langle R, 0 \rangle$, where $R$ is one of the grammar rules. Thus there are 12 items in $I_{predict}$.

$S_1$ contains three more items. Application of $DM$ to the item corresponding to rule 2 in entry $(0,0)$ and to the item corresponding to "John" in $(0,1)$ results in the addition of the edge

$$\begin{bmatrix} \text{word} \\ SYN: & [\text{pn}] \\ HEAD: & \boxed{1} \begin{bmatrix} \text{head} \\ AGR: & \begin{bmatrix} \text{agr} \\ PERS: & [\text{3rd}] \\ NUM: & [\text{sg}] \end{bmatrix} \end{bmatrix} \\ CASE: & \boxed{2}[\text{case}] \end{bmatrix} \bullet \begin{bmatrix} \text{phrase} \\ SYN: & [\text{n}] \\ HEAD: & \boxed{1} \\ CASE: & \boxed{2} \end{bmatrix} \quad (7)$$

to entry $(0, 1)$. In a similar way, the edge

$$\begin{bmatrix} \text{phrase} \\ SYN: & [\mathbf{v}] \\ HEAD: & \boxed{1}\begin{bmatrix} \text{head} \\ AGR: & \begin{bmatrix} \text{agr} \\ NUM: & [\mathbf{sg}] \end{bmatrix} \end{bmatrix} \\ SBCT: & \begin{bmatrix} \text{nelist} \\ 1ST: & \boxed{3} \\ RST: & \boxed{2}[\text{elist}] \end{bmatrix} \end{bmatrix} \bullet \boxed{3}\begin{bmatrix} \text{phrase} \\ SYN: & [\mathbf{n}] \\ HEAD: & [\text{head}] \\ CASE: & [\mathbf{acc}] \end{bmatrix} \begin{bmatrix} \text{phrase} \\ SYN: & [\mathbf{v}] \\ HEAD: & \boxed{1} \\ SBCT: & \boxed{2} \end{bmatrix}$$

(8)

is added to entry $(1, 2)$, by virtue of rule 3 and "loves" and the edge

$$\begin{bmatrix} \text{word} \\ SYN: & [\mathbf{pn}] \\ HEAD: & \boxed{1}\begin{bmatrix} \text{head} \\ AGR: & [\mathbf{agr}] \end{bmatrix} \\ CASE: & \boxed{2}[\mathbf{case}] \end{bmatrix} \bullet \begin{bmatrix} \text{phrase} \\ SYN: & [\mathbf{n}] \\ HEAD: & \boxed{1} \\ CASE: & \boxed{2} \end{bmatrix}$$

(9)

is added to entry $(2, 3)$ by virtue of the rule 2 and "fish".

Several items are added to $S_2$, two of which are of more interest. On the basis of item 7 and the item corresponding to rule 1 in $(0, 0)$, the edge

$$\begin{bmatrix} \text{phrase} \\ SYN: & [\mathbf{n}] \\ HEAD: & \boxed{1}\begin{bmatrix} \text{head} \\ AGR: & \boxed{3}\begin{bmatrix} \text{agr} \\ PERS: & [\mathbf{3rd}] \\ NUM: & [\mathbf{sg}] \end{bmatrix} \end{bmatrix} \\ CASE: & [\mathbf{case}] \end{bmatrix} \bullet \begin{bmatrix} \text{phrase} \\ SYN: & [\mathbf{v}] \\ HEAD: & \boxed{2}\begin{bmatrix} \text{head} \\ AGR: & \boxed{3} \end{bmatrix} \\ SBCT: & [\mathbf{elist}] \end{bmatrix} \begin{bmatrix} \text{phrase} \\ SYN: & [\mathbf{s}] \\ SUBJ: & \boxed{1} \\ HEAD: & \boxed{2} \end{bmatrix}$$

(10)

is added to $(0, 1)$. On the basis of item 8 in $(1, 2)$ and item 9 in $(2, 3)$, the following edge is added to $(1, 3)$:

$$\begin{bmatrix} \text{phrase} \\ SYN: & [\mathbf{v}] \\ HEAD: & \boxed{1}\begin{bmatrix} \text{head} \\ AGR: & \begin{bmatrix} \text{agr} \\ NUM: & [\mathbf{sg}] \end{bmatrix} \end{bmatrix} \\ SBCT: & \begin{bmatrix} \text{nelist} \\ 1ST: & \boxed{3} \\ RST: & \boxed{2}[\text{elist}] \end{bmatrix} \end{bmatrix} \boxed{3}\begin{bmatrix} \text{phrase} \\ SYN: & [\mathbf{n}] \\ HEAD: & \begin{bmatrix} \text{head} \\ AGR: & [\mathbf{agr}] \end{bmatrix} \\ CASE: & [\mathbf{acc}] \end{bmatrix} \bullet \begin{bmatrix} \text{phrase} \\ SYN: & [\mathbf{v}] \\ HEAD: & \boxed{1} \\ SBCT: & \boxed{2} \end{bmatrix}$$

(11)

This complete edge can now be used with edge 10 to form, in $S_3$, the following edge in $(0, 3)$:

$$\begin{bmatrix} \text{phrase} \\ SYN: & [\mathbf{n}] \\ HEAD: & \boxed{1}\begin{bmatrix} \text{head} \\ AGR: & \boxed{3}\begin{bmatrix} \text{agr} \\ PERS: & [\mathbf{3rd}] \\ NUM: & [\mathbf{sg}] \end{bmatrix} \end{bmatrix} \\ CASE: & [\mathbf{case}] \end{bmatrix} \begin{bmatrix} \text{phrase} \\ SYN: & [\mathbf{v}] \\ HEAD: & \boxed{2}\begin{bmatrix} \text{head} \\ AGR: & \boxed{3} \end{bmatrix} \\ SBCT: & [\mathbf{elist}] \end{bmatrix} \bullet \begin{bmatrix} \text{phrase} \\ SYN: & [\mathbf{s}] \\ SUBJ: & \boxed{1} \\ HEAD: & \boxed{2} \end{bmatrix}$$

(12)

and since the head of this complete edge, which spans the entire input string, is more specific than the initial symbol, the string "John loves fish" is accepted by the parser.